\documentclass{article}

\usepackage{PRIMEarxiv}

\usepackage[utf8]{inputenc} 
\usepackage[T1]{fontenc}    
\usepackage{hyperref}       
\usepackage{url}            
\usepackage{booktabs}       
\usepackage{amsfonts}       
\usepackage{nicefrac}       
\usepackage{microtype}      

\usepackage{graphicx} 
\usepackage{float}
\usepackage[version=4]{mhchem} 
\usepackage{amsmath} 
\usepackage{amssymb} 
\usepackage{textcomp,gensymb} 
\usepackage{appendix} 
\usepackage{colortbl} 
\usepackage{rotating} 
\usepackage{bm} 
\usepackage{multirow} 
\usepackage{longtable} 
\usepackage{booktabs} 
\usepackage{multicol} 
\usepackage{multicol}
\usepackage{subcaption}
\usepackage{setspace}
\counterwithin{figure}{section}
\counterwithin{table}{section}

\usepackage{amsmath}
\usepackage{tabularx}
\usepackage{cancel}
\usepackage{hyperref}
\usepackage{tikz}
\usepackage{standalone}
\usepackage{tabularx}
\usepackage{adjustbox}
\usepackage{amssymb}
\usepackage{algorithm, algpseudocode}
\usepackage{graphicx}
\usepackage{cancel}

\usepackage{lipsum}
\usepackage{fancyhdr}       
\usepackage{graphicx}       
\graphicspath{{media/}}     

\title{Stochastic Thermodynamics of Learning Parametric Probabilistic Models}

\author{
  Shervin Sadat Parsi \\
  Physics program at The Graduate Center \\
  City University of New York\\
  \texttt{\{shsparsi\}@gmail.com} \\
}

\begin{document}
\maketitle

\begin{abstract}
We have formulated a family of machine learning problems as the time evolution of Parametric Probabilistic Models (PPMs), inherently rendering a thermodynamic process. Our primary motivation is to leverage the rich toolbox of thermodynamics of information to assess the information-theoretic content of learning a probabilistic model. We first introduce two information-theoretic metrics: Memorized-information (M-info) and Learned-information (L-info), which trace the flow of information during the learning process of PPMs. Then, we demonstrate that the accumulation of L-info during the learning process is associated with entropy production, and parameters serve as a heat reservoir in this process, capturing learned information in the form of M-info. 
\end{abstract}

\tableofcontents 

\section{Introduction}

Starting from nearly half a century ago, physicists began to learn that information is a physical entity \cite{landauer, Szilard, bennett}. Today, the information-theoretic perspective has significantly impacted various fields of physics, including quantum computing \cite{nielsen2010quantum}, cosmology \cite{cosmo}, and thermodynamics \cite{thermo_of_info}. Simultaneously, recent years have witnessed the remarkable success of an algorithmic approach known as machine learning, which is adept at learning information from data. This paper is propelled by a straightforward proposition: if "information is physical", then the process of learning information must inherently be a physical process.

The concepts of memory, prediction, and information exchange between subsystems have undergone extensive exploration within the realms of Thermodynamics of Information \cite{thermo_of_info} and Stochastic Thermodynamics \cite{stoch-thermo}. For instance, Still et al. \cite{still_sivak_bell_crooks_2012} delved into the thermodynamics of prediction. And, the role of information exchange between thermodynamic subsystems has been studied by Sagawa and Ueda \cite{sagawa2012fluctuation}, and Esposito et al.  \cite{EP_corr}. This rich toolbox of thermodynamic of information is our main venue to study physics of machine learning process, with motivation to assess the information content of the learning process. 

The type of machine learning problems we consider in this study encompasses any algorithmic approach that \textit{evolves} a Parametric Probabilistic Model (PPM), or simply the model, towards a desirable target distribution through gradient-based loss function minimization. To establish our notation, consider a set of observations denoted by the training dataset $B$, drawn from an unknown target distribution $p^*$. The PPM, without lose of generality, can be written as the follows:
\begin{equation}\label{generic-ppm}
    p_\theta(X = x) = e^{-\phi_\theta(x)}
\end{equation}
This distribution is parameterized by a set of parameters $\theta \in \mathbb{R}^M$. The objective of learning is to find a set of parameters such that samples drawn from the model, $x \sim p(X|\theta)$, exhibit desirable statistical characteristics. In machine learning practice, one constructs the function $\phi_\theta(x)$ with a (deep) neural network and leave the parameter selection task to an optimizer that minimizes a loss function. Examples encompass Energy-based models \cite{ebm}, Large Language Models, Softmax classifiers, Variational Autoencoders (VAEs) \cite{vae}, among others.

While the information-theoretic approach to this problem is prevalent in the field \cite{jeon2022information, yi2022mutual, shwartzziv2023compress, nn-info}, it has also faced criticisms \cite{cri_info}. Our primary motivation for framing learning in a PPM as a thermodynamic process is to facilitate the assessment of the information content inherent in the learning process. The structure of this paper is outlined as follows: Section 2 briefly discusses prior information-theoretic approaches to the learning problem with PPM, and the challenges they encounter. Subsequently, we introduce our own information-theoretic metrics. Finally, sections 3 and 4 employ the thermodynamic framework to address these information-theoretic inquiries.

\section{Information content of PPMs}\label{sec: info measure}

Locating information within the parametric model, a.k.a. the neural network, remains a fundamental question in machine learning machine \cite{whereisinfo}. This challenge is central to any information-theoretic perspective on machine learning problems. In a pioneering study, Shwartz-Ziv et al. \cite{shwartz2017opening} quantified the internal information within neural networks by estimating the mutual information between inputs and the activities of hidden neurons. Moreover, they employed the information bottleneck theory to interpret the decrease in this mutual information as evidence of data compression during learning. This perspective garnered significant attention in the field \cite{nn-info, shwartzziv2023compress}, reinforcing the view of neural networks as an information channel. However, the study encountered critiques \cite{cir-saxe, cri_info}. A primary problem was that the hidden neurons' activity in a neural network constitutes a deterministic function of the input. Such determinism inherently possesses a trivial mutual information value, even prior to any learning. The challenge of defining a well-defined and interpretable (Shannon) information metric in deterministic neural networks has prompted the proposition that neural network information processing is geometric in nature \cite{cir-geiger} (given that inputs are mapped to a latent space of differing dimensions), rather than information-theoretical.

In a distinct research direction, Ref. \cite{hinton-weight} addresses the significance of assessing the information content of the model's parameters. In our study, we echo this view, emphasizing that parameters are the primary carriers of learned information within neural networks. Consequently, any information-theoretic measure of learned information by the model should be grounded in parameters rather than the deterministic activity of hidden neurons. However, quantifying the information within parameters poses challenges, primarily due to the elusive nature of their distribution \cite{achille2018emergence}. In this section, we introduce two information-theoretic metrics crafted to assess the information content within the learning process of a PPM. This paves the way for computing these quantities within the thermodynamic framework.

To avoid introducing new notation, we also denote $B$ as the ground truth random variable associated with the target distribution $p^*$ from which the training dataset is sampled. Subsequently, we represent the action of the optimizer as a map between this ground truth random variable and the desired set of parameters after $n$ optimization steps:
\begin{equation} \label{map}
    \theta_{t_n} = \Lambda_n(B)
\end{equation}
The map $\Lambda_n$ incorporates the structure of the loss function, the optimization algorithm, and any hyperparameters related to the optimizer's action. We exclude the initial parameters' value from this map's argument, under the assumption that as $n$ increases, the final set of parameters becomes independent of its initial condition. In Information Theory terminology, this map corresponds to a \textit{statistic} of the ground truth random variable \cite{cover}. Moreover, the outcome of this map defines a model, from which the final model-generated sample is sampled: $x_{t_n} \sim p(x|\theta_{t_n})$. Considering that the model-generated sample becomes independent of the ground truth random variable given the parameters, we can express the following Markov chain governing the learning process:
\begin{equation}\label{main markov}
     B \rightarrow{} \theta_{t_n}  \xrightarrow{} x_{t_n}.
\end{equation}
The Data Processing Inequality (DPI) associated with this Markov chain serves as our framework to define two information-theoretic metrics that gauge the information content of the model:
\begin{equation}\label{dpi}
   \underbrace{I_{\Theta;B}(t_n)}_{\text{M-info}}  \geq \underbrace{I_{X;B}(t_n)}_{\text{L-info}}.
\end{equation}
We have used notations presented in table \ref{tab:notations}. The left-hand side of this inequality quantifies the accumulation of mutual information between the parameters and the training dataset, while the right-hand side characterizes the performance of the generative model, it gauges the accumulation of mutual information between the model's generated samples and the training dataset. We refer to the former as Memorized Information (M-info) and the latter as Learned-information (L-info). We also note that both of these quantities start at zero before the training begins. Thus, their measurements at $t_n$, reveal accumulation of information during the learning process. \par

\begin{table}[t]
\centering

\begin{tabularx}{\textwidth}{|lX|}
    \hline
    \\
    $\Delta_{t_n} f(t) := f({t_n}) - f({0})$ & Change over the interval $[0,t_n]$\\[2ex]
    $<f(x)>_{p(x)}:= \int dx ~p(x)~f(x)$ & Average over $p(x)$\\[2ex]
    $s_X(t):= s[p_t(x)]:=  - \ln{p_t(x)}$ & Surprisal of $p_t(x)$\\[2ex]
    $S_X(t):=S[p_t(x)]: = <- \ln{p_t(x)}>_{p_t(x)}$ & Shannon entropy of $p_t(x)$\\[2ex]
    $I_{X;\Theta}(t): = I[X_t;\Theta_t]:= S_X(t) - S_{X|\Theta}(t)$ & Mutual information between $X$ and $\Theta$ at time t\\[2ex]
    \hline
\end{tabularx}%
\vspace{2mm}
\caption{A list of notations used in this paper}
\label{tab:notations}
\end{table}

In the context of the learning problem, the DPI as referenced in \ref{dpi} suggests that what is \textit{Memorized} is always greater than or equal to what is \textit{Learned}. The L-info metric is task-oriented. For example, in the realm of image generation, it quantifies the statistical resemblance between the model's outputs and the genuine images. In the case of classification task, L-info would encapsulate only the pertinent information for label prediction. In contrast, M-info can encompass information not directly pertinent to the current task. For instance, it might capture intricate pixel configurations in an image dataset, which aren't crucial for identifying distinct patterns like human faces. The DPI \ref{dpi} neatly illustrate the risk of overfitting, when a model starts to incorporate extraneous information that doesn't align with the learning objective. The necessity of constraining the information in a model's parameters is highlighted in Ref. \cite{hinton-weight}, echoing the Minimum Description Length Principle \cite{Rissanen1986StochasticCA}. Additionally, studies suggest that the SGD optimizer tends to favor models with minimal information in their parameters \cite{achille2018emergence}. Recent work by Ref. \cite{bu2020tightening} has even proposed an upper limit for minimizing parameter information to bolster generalization capabilities. These findings suggest that the learning process seeks to minimize the left-hand side of the DPI inequality while simultaneously maximizing the right-hand side, that measures the model performance. This leads us to an ideal scenario where $I_{\Theta;B}(t_n) = I_{X;B}(t_n)$, signifying that all memorized information is relevant to the learning task.

We now take one step further in our definition of M-info and L-info. First, the presence of the optimizer map, as referenced in \ref{map}, connecting the ground truth source of the training dataset to the parameters, allows us to simplify M-info as follows:
\begin{equation}\label{eq: M-info}
    \text{M-info}: = I_{\Theta;B}(t_n)  = S(\Theta_{t_n})
\end{equation}
Thus, the parameters naturally emerge as the model's \textit{memory}, where its Shannon entropy measures the stored information during the learning process. \par

Second, we swap $B$ for $\Theta_{t_n}$, in the definition of L-info in cost of losing some information: 
\begin{equation}\label{sufficent}
\begin{split}
      \text{L-info}:= I_{X;B}(t_n) &= I_{X;\Lambda_{t_n}(B)} + \epsilon\\
     & = I_{X;\Theta}(t_n)+ \epsilon
\end{split} 
\end{equation}
where $\epsilon$ is a non-negative number that equals zero only when the map $\Lambda_{n}$ outcome is a \textit{sufficient statistic} for $B$. For the above expression, the condition of sufficient statistic can be eased as $\Theta$ to be sufficient with respect to $X$. This means the map $\Lambda_{n}$ preserve all information in $B$ that is also mutual in $X$. Indeed, in the problem, we are interested in this type of preservative maps that their action on training dataset preserve task-related information. Therefore, we consider $I_{X;\Theta}$ as a reasonable proxy to L-info, and we use the two interchangeably:
\begin{equation}\label{eq: L-info}
    \text{L-info}: = I_{X;\Theta}(t_n)
\end{equation}

\section{The learning trajectory of a PPM}\label{sec: Fundamental Concepts}
\label{sec: the learning trajectory}

\begin{figure}\centering
\tikzset{every picture/.style={line width=0.75pt}} 

\begin{tikzpicture}[x=0.75pt,y=0.75pt,yscale=-1,xscale=1]

\draw  [fill={rgb, 255:red, 208; green, 2; blue, 27 }  ,fill opacity=0.27 ] (289.6,98.9) .. controls (289.6,85.92) and (300.12,75.4) .. (313.1,75.4) .. controls (326.08,75.4) and (336.6,85.92) .. (336.6,98.9) .. controls (336.6,111.88) and (326.08,122.4) .. (313.1,122.4) .. controls (300.12,122.4) and (289.6,111.88) .. (289.6,98.9) -- cycle ;
\draw    (139,168.4) .. controls (173.65,134.74) and (237.7,153.03) .. (266.15,108.76) ;
\draw [shift={(267,107.4)}, rotate = 121.33] [color={rgb, 255:red, 0; green, 0; blue, 0 }  ][line width=0.75]    (10.93,-3.29) .. controls (6.95,-1.4) and (3.31,-0.3) .. (0,0) .. controls (3.31,0.3) and (6.95,1.4) .. (10.93,3.29)   ;
\draw  [fill={rgb, 255:red, 126; green, 211; blue, 33 }  ,fill opacity=0.2 ] (98,67.5) -- (361,67.5) -- (361,194.5) -- (98,194.5) -- cycle ;

\draw (117,166.4) node [anchor=north west][inner sep=0.75pt]    {$p_{\theta }{}_{0} \ $};
\draw (307,86.4) node [anchor=north west][inner sep=0.75pt]    {$p^{*}$};
\draw (256,83.4) node [anchor=north west][inner sep=0.75pt]    {$p_{\theta }{}_{n} \ $};
\draw (181,120.4) node [anchor=north west][inner sep=0.75pt]    {$\mathcal{T}$};

\end{tikzpicture}
\caption[The Learning Trajectory]{The learning trajectory $\mathcal{T}$ depicts the thermodynamic process that take the initial model state to final state. The green area shows the space of family of distribution accessible to the PPM. The red area considers the possibility that the target distribution, $p^*$, is not in this family.} \label{fig: trajectory}
\end{figure}
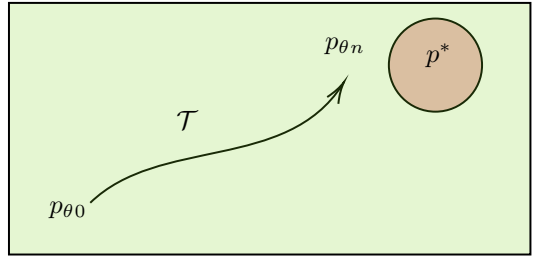

The time evolution of the PPM is the first clue to frame the learning process as a thermodynamic process. To illustrate this, consider a discretized time interval $[0, t_n]$, which represents the time needed for $n$ optimization steps of the parameters. During this time, the optimizer algorithm draws a sequence of i.i.d samples from the training dataset. We denote this sequence by $\boldsymbol{b}_{n} : = \{b_{t_1}, b_{t_2}, \dots, b_{t_n}\}$, and refer to it as the "input trajectory". Then, the outcome of the optimization defines a sequence of parameters, call it the "parameters' trajectory": $\boldsymbol{\theta}_{n} := \{\theta_0, \theta_{t_1}, \theta_{t_2}, \dots, \theta_{t_n}\}$. Each realization of parameters defines a specific PPM. Consequently, the parameters' trajectory produces a sequence of PPMs:
\begin{equation}\label{eq: learning-trajectory}
\mathcal{T} := \{p(X|\theta_{0}),~p(X|\theta_{t_1}),~p(X|\theta_{t_2}), \dots, ~p(X|\theta_{t_n})\}
\end{equation}
We refer to this sequence as the \textit{learning trajectory}, depicted in figure \ref{fig: trajectory}. On the other hand, a thermodynamic process can be constructed solely from the time evolution of a distribution \cite{esposito_intro}. Therefore, we see $\mathcal{T}$ as a thermodynamic process. The physics of this process is encoded in the transition rates governing the master equation of this time evolution. Finding the transition rate associated to learning a PPM, is our main task in this section.  

\subsection{The model subsystem}

We refer to the subsystem that goes under the thermodynamic process $\mathcal{T}$ as \textit{the model subsystem}. This subsystem has $X$ degrees of freedom, and its microscopic states' realization along the learning trajectory represent model-generated samples at each time step: $ x_{t_i} \sim p(X|\theta_{t_i})$. Furthermore, we denote the stochastic trajectory of model-generated samples by $\boldsymbol{x}_{n} := \{x_{t_1}, x_{t_2}, \dots, x_{t_n}\}$. To avoid confusion with our notation, consider the probability functions $p(x_{t_i}|\theta_{t_i})$ and $p(x_{t_{i-1}}|\theta_{t_i})$, which respectively represent the probability of observing $x_{t_i} \in \boldsymbol{x}_{n}$ and $x_{t_{i-1}} \in \boldsymbol{x}_{n}$ at time $t=t_i$. Here, the time index of $\theta$ aligns with the time index of the PPM, i.e., $p_{t_i}(X|\theta_{t_i})\equiv p(X|\theta_{t_i}) $, because the PPM is fully defined upon observing the parameters. In contrast, the time index on $x$ denotes a specific observation within $\boldsymbol{x}_n$. To simplify our notation, the absence of a time index on $x$ denotes a generic realization of the random variable $X$, and we write $p(x|\theta_{t_i})$ instead of $p(X|\theta_{t_i})$.

\subsection{The parameters subsystem} \label{ensemble view sub}

The parameters of the neural network at each step of optimization represent realization of the parameters subsystem, with $\Theta$ degrees of freedom, and the stochastic trajectory $\boldsymbol{\theta}_{n}$. The statistical state of the parameters subsystem is given with the marginal $p(\theta_{t_i})$ at time step $t= t_i$. This marginal state represents the statistic of all possible outcome of training a PPM on specific learning objective. We can think of training an ensemble of computers on the same machine learning task. This allows us to think about the time evolution of the marginal $p(\theta_{t_i})$, and the joint distribution $p(x|\theta_{t_i}) p(\theta_{t_i})$ during the learning process. we refer to this view as the \textit{ensemble view} of learning process. In practice, however, we train the PPM only once, and we do not have access to the marginal $p(\theta_{t_i})$. Thus, our model-generated samples are conditioned on specific observations of parameters, $\theta_{t_i} \sim \Theta_{t_i}$. This defines the \textit{conditional view} of the learning process, that is fully described by the learning trajectory of the PPM.

In machine learning practice, it is desirable for a training process to exhibit a robust outcome, regardless of who is running the code. One way to achieve this is by imposing a \textit{low-variance condition} on the statistics of parameters across the ensemble of all learning trials. This condition asserts that the parameters' trajectory across the ensemble is confined to a small region $D_n \subset \mathcal{R}^M$. As $n$ grows larger, this region shrinks, and becomes associated with the area surrounding the target distribution as depicted in Figure \ref{fig: trajectory}. Under this condition, we can express:
\begin{equation}\label{low-var condtion}
    <f(\theta)>_{p(\theta_n)} \approx f(\theta^*), ~~\forall \theta^* \in D_n 
\end{equation}
The above approximation becomes exact when $p(\theta_n)$ assumes the form of a delta-Dirac function, indicating a zero-variance condition in the parameters' dynamics.

The low-variance condition proves invaluable when computing the information-theoretic measurements introduced in section \ref{sec: info measure}. This is because the computation of the M-info $I_{B;\Theta}$ and L-info $I_{X;\Theta}$ necessitates averaging over the parameters' distribution. However, since we typically train our model just once, we lack direct access to the parameters' distribution throughout the learning trajectory. To overcome this challenge, we introduce the Conditional L-info:
\begin{equation} \label{conditioned l-info}
I_{X:\Theta}(\theta_t)= \int dx ~ p(x|\theta_t) \ln{\frac{ p(x|\theta_t)}{ p_t(x)}} 
\end{equation}
Subsequently, under the low-variance condition of the parameters subsystem, we can measure the conditional L-info as a proxy for the L-info: $I_{X:\Theta}(t) = <I_{X:\Theta}(\theta)>_{p_t(\theta)} \approx I_{X:\Theta}(\theta_t) $.
In section \ref{sub par dynamic}, we will delve deeper into the evidence supporting the low-variance condition of the subsystem $\Theta$.

We refer to the joint $(X,\Theta)$ as the \textit{learning system}, that embodies the thermodynamic process of learning a PPM. In this section, we will demonstrate that the thermodynamic exchange between model subsystem and parameters subsystem is the primary source producing M-info and L-info during the learning process. Before delving further, we establish two interconnected assumptions about the parameters subsystem: 
(1) The PPM is over-parameterized; specifically, the subsystem $\Theta$ has a much higher dimension compared to the subsystem $X$.  
(2) The parameters subsystem evolves in a quasi-static fashion (slow dynamics).

The foundation for these assumptions in machine learning is clear. Training over-parameterized models represents a pivotal achievement of machine learning algorithms, and the slow dynamics (often termed as lazy dynamics) of these over-parametrized models are well-documented \cite{lazy1,lazy2}. These characteristics underscore the significant role of the parameters subsystem in the learning process, akin to that of a heat reservoir. Over-parameterization implies a higher heat capacity for this subsystem compared to the model subsystem. Additionally, the quasi-dynamics align with the behavior of an ideal heat reservoir, which doesn't contribute to entropy production \cite{deffner2013information}. The role of the parameters subsystem as a reservoir aligns with the assumption of a low-variance condition for this subsystem. This is because we expect the stochastic dynamics of a reservoir in contact with the subsystem to be low-variance across the ensemble of all trials. 

In this study, we attribute the role of an ideal heat reservoir to the parameters subsystem, with inverse temperature $\beta^{-1} = 1$. In section \ref{sub par dynamic}, we delve deeper into the rationale behind this assumption, by examining the stochastic dynamics of parameters under a vanilla stochastic gradient descent optimizer, and highlighting potential limitations of this assumption.

\subsubsection{Lagged bipartite dynamics}\label{biparti and timescale subsec}
We want to emphasize that the dynamics of subsystem $X$ is not a mere conjecture or an arbitrary component in this study; rather, it's an integral part of training a generative PPM. This dynamics is inherent in the optimizer action, necessitating a fresh set of model-generated samples to compute the loss function or its gradients after each parameter update. For instance, in the context of EBM, a Langevin Monte Carlo (LMC) sampler can be employed to generate new samples from the model \cite{du2019implicit}. The computational cost of producing a fresh set of model-generated samples introduces a time delay in the parameter dynamics. For instance, when using an LMC sampler, the number of Monte Carlo steps dictates this lag time. Conversely, in the case of a language model, since the computation of the loss function relies on inferring subsequent tokens, the inference latency signifgies the time delay.
\begin{figure}\centering
\tikzset{every picture/.style={line width=0.75pt}} 

\begin{tikzpicture}[x=0.75pt,y=0.75pt,yscale=-1,xscale=1]

\draw    (54.17,84) -- (105.61,83.86) ;
\draw [shift={(107.61,83.86)}, rotate = 179.85] [color={rgb, 255:red, 0; green, 0; blue, 0 }  ][line width=0.75]    (10.93,-3.29) .. controls (6.95,-1.4) and (3.31,-0.3) .. (0,0) .. controls (3.31,0.3) and (6.95,1.4) .. (10.93,3.29)   ;
\draw    (133.5,84.67) -- (184.94,84.53) ;
\draw [shift={(186.94,84.52)}, rotate = 179.85] [color={rgb, 255:red, 0; green, 0; blue, 0 }  ][line width=0.75]    (10.93,-3.29) .. controls (6.95,-1.4) and (3.31,-0.3) .. (0,0) .. controls (3.31,0.3) and (6.95,1.4) .. (10.93,3.29)   ;
\draw  [dash pattern={on 4.5pt off 4.5pt}]  (54.61,129.19) -- (103.28,129.19) ;
\draw [shift={(105.28,129.19)}, rotate = 180] [color={rgb, 255:red, 0; green, 0; blue, 0 }  ][line width=0.75]    (10.93,-3.29) .. controls (6.95,-1.4) and (3.31,-0.3) .. (0,0) .. controls (3.31,0.3) and (6.95,1.4) .. (10.93,3.29)   ;
\draw  [dash pattern={on 4.5pt off 4.5pt}]  (135.61,130.19) -- (184.28,130.19) ;
\draw [shift={(186.28,130.19)}, rotate = 180] [color={rgb, 255:red, 0; green, 0; blue, 0 }  ][line width=0.75]    (10.93,-3.29) .. controls (6.95,-1.4) and (3.31,-0.3) .. (0,0) .. controls (3.31,0.3) and (6.95,1.4) .. (10.93,3.29)   ;
\draw  [color={rgb, 255:red, 208; green, 2; blue, 27 }  ,draw opacity=1 ] (22,179.18) -- (428,179.18) -- (428,219.18) -- (22,219.18) -- cycle ;
\draw [color={rgb, 255:red, 208; green, 2; blue, 27 }  ,draw opacity=1 ]   (158.88,159.26) -- (159.49,168.92)(155.89,159.44) -- (156.5,169.11) ;
\draw [shift={(158.5,177)}, rotate = 266.38] [color={rgb, 255:red, 208; green, 2; blue, 27 }  ,draw opacity=1 ][line width=0.75]    (10.93,-3.29) .. controls (6.95,-1.4) and (3.31,-0.3) .. (0,0) .. controls (3.31,0.3) and (6.95,1.4) .. (10.93,3.29)   ;
\draw    (99.17,198) -- (119.88,198.32) ;
\draw [shift={(121.88,198.35)}, rotate = 180.88] [color={rgb, 255:red, 0; green, 0; blue, 0 }  ][line width=0.75]    (10.93,-3.29) .. controls (6.95,-1.4) and (3.31,-0.3) .. (0,0) .. controls (3.31,0.3) and (6.95,1.4) .. (10.93,3.29)   ;
\draw    (58.48,93.84) -- (100.07,117.52) ;
\draw [shift={(101.81,118.51)}, rotate = 209.65] [color={rgb, 255:red, 0; green, 0; blue, 0 }  ][line width=0.75]    (10.93,-3.29) .. controls (6.95,-1.4) and (3.31,-0.3) .. (0,0) .. controls (3.31,0.3) and (6.95,1.4) .. (10.93,3.29)   ;
\draw    (217.5,85.67) -- (268.94,85.53) ;
\draw [shift={(270.94,85.52)}, rotate = 179.85] [color={rgb, 255:red, 0; green, 0; blue, 0 }  ][line width=0.75]    (10.93,-3.29) .. controls (6.95,-1.4) and (3.31,-0.3) .. (0,0) .. controls (3.31,0.3) and (6.95,1.4) .. (10.93,3.29)   ;
\draw  [dash pattern={on 4.5pt off 4.5pt}]  (219.61,131.19) -- (268.28,131.19) ;
\draw [shift={(270.28,131.19)}, rotate = 180] [color={rgb, 255:red, 0; green, 0; blue, 0 }  ][line width=0.75]    (10.93,-3.29) .. controls (6.95,-1.4) and (3.31,-0.3) .. (0,0) .. controls (3.31,0.3) and (6.95,1.4) .. (10.93,3.29)   ;
\draw [color={rgb, 255:red, 0; green, 0; blue, 0 }  ,draw opacity=1 ][fill={rgb, 255:red, 0; green, 0; blue, 0 }  ,fill opacity=1 ]   (140.81,95.84) -- (182.41,119.52) ;
\draw [shift={(184.14,120.51)}, rotate = 209.65] [color={rgb, 255:red, 0; green, 0; blue, 0 }  ,draw opacity=1 ][line width=0.75]    (10.93,-3.29) .. controls (6.95,-1.4) and (3.31,-0.3) .. (0,0) .. controls (3.31,0.3) and (6.95,1.4) .. (10.93,3.29)   ;
\draw    (221.98,96.34) -- (263.57,120.02) ;
\draw [shift={(265.31,121.01)}, rotate = 209.65] [color={rgb, 255:red, 0; green, 0; blue, 0 }  ][line width=0.75]    (10.93,-3.29) .. controls (6.95,-1.4) and (3.31,-0.3) .. (0,0) .. controls (3.31,0.3) and (6.95,1.4) .. (10.93,3.29)   ;
\draw  [draw opacity=0][fill={rgb, 255:red, 207; green, 95; blue, 95 }  ,fill opacity=0.25 ] (101,107) -- (215,107) -- (215,157.5) -- (101,157.5) -- cycle ;
\draw    (246.17,199) -- (266.88,199.32) ;
\draw [shift={(268.88,199.35)}, rotate = 180.88] [color={rgb, 255:red, 0; green, 0; blue, 0 }  ][line width=0.75]    (10.93,-3.29) .. controls (6.95,-1.4) and (3.31,-0.3) .. (0,0) .. controls (3.31,0.3) and (6.95,1.4) .. (10.93,3.29)   ;

\draw (109.33,118.4) node [anchor=north west][inner sep=0.75pt]    {$x_{t_{0}} \ $};
\draw (33.33,73.4) node [anchor=north west][inner sep=0.75pt]    {$\theta _{0\ } \ $};
\draw (111.67,73.4) node [anchor=north west][inner sep=0.75pt]  [color={rgb, 255:red, 74; green, 144; blue, 226 }  ,opacity=1 ]  {$\theta _{t_{1}} \ $};
\draw (191.67,74.07) node [anchor=north west][inner sep=0.75pt]    {$\theta _{t_{2}} \ $};
\draw (28.83,187.4) node [anchor=north west][inner sep=0.75pt]    {$( \theta _{t_{1}} ,x_{t_{0} \ }) \ $};
\draw (313,159) node [anchor=north west][inner sep=0.75pt]   [align=left] {{\small \textit{relaxation time }}};
\draw (275.67,122.07) node [anchor=north west][inner sep=0.75pt]    {$x_{t_{2}} \ \ \ ...$};
\draw (275.67,75.07) node [anchor=north west][inner sep=0.75pt]    {$\theta _{t_{3}} \ \ \ ...\ $};
\draw (191.17,118.57) node [anchor=north west][inner sep=0.75pt]  [color={rgb, 255:red, 0; green, 0; blue, 0 }  ,opacity=1 ]  {$x_{t_{1}} \ $};
\draw (19.33,120.4) node [anchor=north west][inner sep=0.75pt]  [font=\footnotesize]  {$x_{rand} \ $};
\draw (111.67,73.4) node [anchor=north west][inner sep=0.75pt]  [color={rgb, 255:red, 0; green, 0; blue, 0 }  ,opacity=1 ]  {$\theta _{t_{1}} \ $};
\draw (125.33,186.39) node [anchor=north west][inner sep=0.75pt]    {$( \theta _{t_{1}} ,x_{t_{0} +\delta t\ }) \ \ \ ...\ \ \ $};
\draw (286.33,188.39) node [anchor=north west][inner sep=0.75pt]    {$( \theta _{t_{1}} ,x_{t_{0} +\tau \delta t\ } \equiv x_{t_{1} \ }) \ $};

\end{tikzpicture}
\caption[Bayesian network of the Joint $(X,\Theta)$]{This figure shows Bayesian network for joint trajectory probability $P[\boldsymbol{x}_n, \boldsymbol{\theta}_n ]$, based on a dual timescale bipartite dynamics.} \label{fig: BN}
\end{figure}
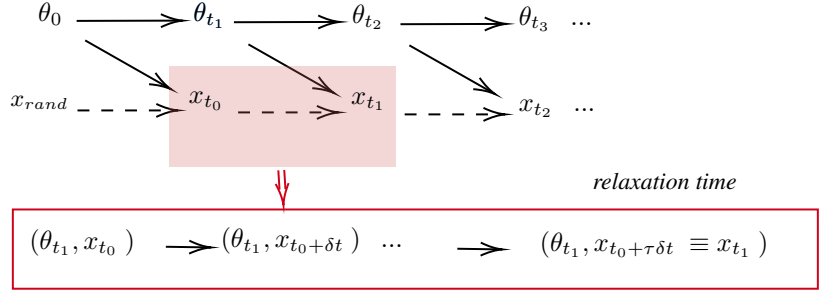
We denote the lag time parameters with $\tau$. Here, the model subsystem evolves on the timescale $\delta t$, while the parameters subsystem evolves on the timescale $\alpha = \tau \delta t$. In the thermodynamic context, this parameter represents the \textit{relaxation time} of the subsystem $X$, under fix microscopic state of subsystem $\Theta$. Conceptually, parameter $\tau$ acts as a complexity metric, quantifying the computational resources required for each parameter optimization step. Moreover, the dynamics of the joint $(X,\Theta)$ exhibit a bipartite property. This implies that simultaneous transitions in the states of $X$ and $\Theta$ are not allowed, given that the observation of a new set of model-generated samples occurs only after a parameter update.

The lagged bipartite dynamics described above can be represented using two time resolutions: $\delta t$ and $\alpha$. In the finer time resolution of $\delta t$, the Markov chain within the time interval $[t_{i}, t_{i+1}]$ is as follows:
\begin{align}\label{high mc}
    (x_{t_{i}} ,\theta_{t_{i}} ) \rightarrow (x_{t_{i}} ,\theta_{t_{i+1}} ) \rightarrow (x_{t_{i}+\delta t} ,\theta_{t_{i+1}} ) ~\dots ~\rightarrow (x_{t_{i}+\tau\delta t} ,\theta_{t_{i+1}} ) \equiv (x_{t_{i+1}} ,\theta_{t_{i+1}} ).
\end{align}
We can also analyze this dynamics at a coarser time resolution of $\alpha$. Within the interval $[t_{0}, t_{n}]$, the Markov Chain appears as:
\begin{align} \label{low mc}
    (x_{0} ,\theta_{0} ) \dashrightarrow (x_{t_{1}} ,\theta_{t_{1}} ) ~\dots ~  (x_{t_{n-1}} ,\theta_{t_{n-1}} ) \dashrightarrow (x_{t_{n}} ,\theta_{t_{n}} ).
\end{align}
In the above Markov chain, the dashed arrows remained us the \textit{ignorance} of intermediate steps in the high resolution picture \ref{high mc}.  Figure \ref{fig: BN}, illustrates the lagged bipartite dynamics of the learning system. An important observation is that the learning trajectory $\mathcal{T}$, as defined in \ref{eq: learning-trajectory}, is written in the low resolution picture. Therefore, studying the learning trajectory means studying the dynamics of the system $(X,\Theta)$ in the low resolution picture.

\subsection{Trajectory probabilities}

To set the stage for the application of the Fluctuation Theorem (FT) to learning a PPM, we define the trajectory probability of the joint $(\boldsymbol{x}n,\boldsymbol{\theta}n)$ as the probability of observing a series of model-generated samples and parameters during the learning process:
\begin{align}\label{ensemble view}
P[\boldsymbol{x}_n,\boldsymbol{\theta}_n] := p(x_0, x_{t_1}, \dots, x_{t_n}, \theta_0, \theta_{t_1}, \dots, \theta_{t_n})
\end{align}
Additionally, we can consider the time reversal of the samples' trajectory and parameters' trajectory, respectively, as $\boldsymbol{\tilde{x}}_n : = \{x_{t_n}, x_{t_{n-1}}, \dots, x_{t_1}\} $ and $\boldsymbol{\tilde{\theta}}_n : = \{\theta_{t_n}, \theta_{t_{n-1}}, \dots, \theta_{t_1}\} $. Then, the probability of observing the backward trajectory is denoted by $P[\boldsymbol{\tilde{x}}_n , \boldsymbol{\tilde{\theta}}_n]$.\par

Here, $P[\boldsymbol{x}_n,\boldsymbol{\theta}_n]$ represents the trajectory probability of the learning system in the ensemble view. In practice, however, we typically train our model only once, and we often lack access to the parameters' distribution. Therefore, our model is conditioned on the observation of a specific parameters' trajectory $\boldsymbol{\theta}_n$. This defines the trajectory probability in the conditional view:
\begin{equation} \label{trajector view}
    P[\boldsymbol{x}_n| \boldsymbol{\theta}_n ] := \frac{P[\boldsymbol{x}_n,\boldsymbol{\theta}_n]}{P[\boldsymbol{\theta}_n]}
\end{equation}
where, 
\begin{equation}
    P[\boldsymbol{\theta}_n] = p( \theta_0, \theta_{t_1}, \dots, \theta_{t_n}). 
\end{equation}

Similarly, the backward conditional trajectory probability is the probability of observing the time-reversal samples' trajectory, conditioned on observation of the time-reversal parameters' trajectory: $P[\boldsymbol{\tilde{x}}_n | \boldsymbol{\tilde{\theta}}_n] = \frac{P[\boldsymbol{\tilde{x}}_n , \boldsymbol{\tilde{\theta}}_n]}{P[ \boldsymbol{\tilde{\theta}}_n]} $.

We now use the Markov property in the Markov chains \ref{high mc} and \ref{low mc} respectively, to decompose the conditional trajectory probability and the marginal trajectory probability as fallows: 
\begin{equation} \label{traj prob decompostion}
\begin{split}
     P[\boldsymbol{x}_n|\boldsymbol{\theta}_n] & = p(x_{t_n}|x_{t_{n-1}}, \theta_{t_{n}})\dots p(x_{t_{1}}|x_{0}, \theta_{t_{1}})  p(x_{0}|\theta_{0}), \\
     P[\boldsymbol{\theta}_n] & = p(\theta_{t_{n}}|\theta_{t_{n-1}})\dots p(\theta_{t_{1}}|\theta_{t_{0}})  p(\theta_{0}), 
\end{split}
\end{equation}
where the expressions such as $p(x_{t_n}|x_{t_{n-1}}, \theta_{t_{n}})$ and $p(\theta_{t_{n}}|\theta_{t_{n-1}})$ represent the transition probabilities that determine the probability of moving from one microscopic state to another. Additionally, we define two probability trajectories, conditioned on the initial conditions, which will be used later in the formulation of FT:
\begin{equation} \label{exclude start and end}
\begin{split}
     P[(\boldsymbol{x}_n|\boldsymbol{\theta}_n) | (x_{0}|\theta_{0})] &: = P[(\boldsymbol{x}_n|\boldsymbol{\theta}_n)] / p(x_{0}|\theta_{0}), \\
     P[\boldsymbol{\theta}_n|\theta_0] & := P[\boldsymbol{\theta}_n|\theta_{0}]/  p(\theta_{0}).
\end{split}
\end{equation}
\vspace{2pt}
\subsection{Local Detailed Balance (LDB) for learning PPMs} 

The transition probabilities, represented in \ref{traj prob decompostion}, capture the physics of the learning problem.
Considering a Markov property(i.e., memoryless process) for time evolution of the model subsystem, the transition rate for this subsystem get reduced to PPM:\begin{equation}\label{forward trans prob}
    p(x_{t_i}|x_{t_{i-1}}, \theta_{t_{i}}) = p(x_{t_i}|\theta_{t_{i}}). 
\end{equation}

The above expression suggests that the transition rate between two microscopic states $x_{t_{i-1}}$ and $x_{t_i}$ under the fixed $\theta_{t_{i}}$, to be equivalent to probability of observing $x_{t_i}$ by the PPM itself at $t=t_i$. To reiterate, this is the Markov property that suggests the element inside $\boldsymbol{x}_n$, are independently and freshly drawn from the PPM specified with given parameters along the learning trajectory $\mathcal{T}$. This is especially true where $\tau >>1$. We can generalize this observation for the backward transition probability  $p(x_{t_{i-1}}|x_{t_{i}}, \theta_{t_{i}})$, that represent probability of the backward transition  $(x_{t_{i}} ,\theta_{t_{i}} ) \dashrightarrow (x_{t_{i-1}} ,\theta_{t_{i}} )$ under fixed $\theta_{i}$, as follows: 
\begin{equation}\label{backward trans prob}
    p(x_{t_{i-1}}|x_{t_{i}}, \theta_{t_{i}}) = p(x_{t_{i-1}}|\theta_{t_{i}}). 
\end{equation}
The above expression tells us that the probability of backward transition is equivalent with the probability of observing the sample generated at $t=t_{i-1}$ in $\boldsymbol{x}_n$ with the PPM at time  $t=t_{i}$.\par

Finally, we write the log ratio of forward and backward transitions: 
\begin{equation} \label{LDB}
    \ln{\frac{p(x_{t_i}|x_{t_{i-1}}, \theta_{t_{i}}) }{ p(x_{t_{i-1}}|x_{t_{i}}, \theta_{t_{i}})}} = \ln{\frac{p(x_{t_i}|\theta_{t_{i}})}{p(x_{t_{i-1}}|\theta_{t_{i}})}} = - \Big( \phi_{\theta_{t_i}} ( x_{t_i}) - \phi_{\theta_{t_i}} ( x_{t_{i-1}})\Big), 
\end{equation}
where the second equality is due to Eq. \ref{generic-ppm}. The above expression resembles the celebrated Local Detailed Balance (LDB) \cite{LBD} that relates the log ratio of forward and backward transition probabilities to the difference in potential energy of initial and final state in the transition. The heat reservoir that supports the legitimacy of the above LBD expression for learning PPM is the parameters subsystem, whose temperature has been set to one, as we will discuss it in more details in section \ref{sub par dynamic}. We emphasize that the above LBD has emerged naturally under assumption of the Markov property and a relaxation time for learning a generic generative PPM. It is also important to note that the above LBD is only valid in the low resolution picture.  \par

The LBD relation, presented in Eq. \ref{LDB}, has a profound consequence. It allows us to write the forward conditional probability trajectory, $ P[\boldsymbol{x}_n|\boldsymbol{\theta}_n]$, and the backward conditional probability trajectory, $P[\boldsymbol{\tilde{x}}_n | \boldsymbol{\tilde{\theta}}_n]$, solely based on the series of PPMs in the learning trajectory $\mathcal{T}$:
\begin{equation} \label{eq: set back-forth}
\begin{split}
    P[\boldsymbol{x}_n|\boldsymbol{\theta}_n] &=  p(x_{t_n}| \theta_{t_n})~\dots ~p(x_{t_1}| \theta_{t_1}) ~ p(x_0|\theta_0) \\
    P[\boldsymbol{\tilde{x}}_n | \boldsymbol{\tilde{\theta}}_n]&=  p(x_{0}| \theta_{t_1})~\dots ~p(x_{t_{n-1}}| \theta_{t_n}) ~ p(x_{t_n}| \theta_{t_n})
\end{split}
\end{equation}
This is significant because it renders the application of the FT framework to the learning PPMs practical, as we have access to elements of the learning trajectory. 
\subsection{L-info from fluctuation theorem} \label{sec: L-info from FT}
The version of the fluctuation theorem we are about to apply to the learning PPMs is known as the Detailed Fluctuation Theorem (DFT)\cite{rao2018detailed}. We also note that the machinery we are about to present for measuring information flow in PPMs has been developed to study information exchange between thermodynamic subsystems \cite{sagawa2012fluctuation}. The novelty here lies merely in the application of this machinery to the learning process of a PPM. In this section, we extensively use notations presented in table \ref{tab:notations}. Also, note that the temperature of the parametric reservoir is set to one. Applying DFT in the conditional view, i.e., the conditional forward and backward trajectories defined in Eq. \ref{eq: set back-forth}, results in:
 \begin{equation}\label{conditinal dft}
 \begin{split}
     \sigma_{\boldsymbol{x}_n|\boldsymbol{\theta}_n}& = \ln{\frac{P[\boldsymbol{x}_n|\boldsymbol{\theta}_n ]}{P[\tilde{\boldsymbol{x}}_n|\tilde{\boldsymbol{\theta}}_n ]}} \\
     &= \ln{\frac{P[(\boldsymbol{x}_n|\boldsymbol{\theta}_n) | (x_{0}|\theta_{0})]}{P[(\tilde{\boldsymbol{x}}_n|\tilde{\boldsymbol{\theta}}_n) | (x_{t_n}|\theta_{t_n})]}} + \ln{\frac{p(x_{0}|\theta_{0})}{p(x_{t_n}|\theta_{t_n})}}\\
     &= - q_{\boldsymbol{x}_n}(\boldsymbol{\theta}_n) + s[p(x_{t_n}|\theta_{t_n})] - s[p(x_{t_0}|\theta_{t_0})]
 \end{split}
 \end{equation}
 The first line is due to DFT, which defines the stochastic EP to be the logarithm of the ratio of the forward and backward trajectory probabilities.  The second line is due to the decomposition presented in Eq. \ref{exclude start and end}. Finally, the third line is the consequence of LDB relation \ref{LDB }, and the definition of the stochastic heat flow $ q_{\boldsymbol{x}_n}(\boldsymbol{\theta}_n)$, as the change in the energy of the subsystem $X$ due to alterations in its microscopic state configuration:
 \begin{equation}\label{x heat stoch}
     q_{\boldsymbol{x}_n}(\boldsymbol{\theta}_n):=-\ln{\frac{P[(\boldsymbol{x}_n|\boldsymbol{\theta}_n) | (x_{0}|\theta_{0})]}{P[(\tilde{\boldsymbol{x}}_n|\tilde{\boldsymbol{\theta}}_n) | (x_{t_n}|\theta_{t_n})]}} = \sum^n_{i=1}  ~ \phi_{\theta_{t_i}}(x_i) -\phi_{\theta_{t_i}}(x_{i-1}).
 \end{equation}
Note that our sing convention defines $ q_{\boldsymbol{x}_n}>0$ as the heat observed by the subsystem $X$. \par

The second law arises from averaging Eq.\ref{conditinal dft} over the forward trajectory distribution $P_F[\boldsymbol{x}_n|\boldsymbol{\theta}_n ]$, and recalling the non-negativity property of the Kl-divergence to establish non-negativity of averaged EP: $\Sigma_{X|\Theta}(\boldsymbol{\theta}_n): = <\ln{\frac{P_F[\boldsymbol{x}_n|\boldsymbol{\theta}_n ]}{P_B[\tilde{\boldsymbol{x}}_n|\tilde{\boldsymbol{\theta}}_n ]}}>_{P_F[\boldsymbol{x}_n|\boldsymbol{\theta}_n ]} ~ \geq 0 $. We note that the averaged EP is still conditioned on the stochastic trajectory of parameters, thus we refer to this as the conditional EP. This is indeed the consequence of working in the conditional view.\par

Motivated to compute L-info, in the next step, we rearrange Eq. \ref{conditinal dft} as follows:
 \begin{equation}\label{stoch l-info}
     \mathcal{I}[x_{t_n}:\theta_{t_n}] - \mathcal{I}[x_{0}:\theta_{0}] = - q_{\boldsymbol{x}_n}(\boldsymbol{\theta}_n) + s[p(x_{t_n})] - s[p(x_{t_0})] - \sigma_{\boldsymbol{x}_n|\boldsymbol{\theta}_n},
  \end{equation}
where  $\mathcal{I}[x_{t_n}:\theta_{t_n}] := s[p(x_{t_n})] - s[p(x_{t_n}|\theta_{t_n})]$ is the mutual content (or stochastic mutual information) at $t=t_n$.  We now arrive at the conditional L-info \ref{conditioned l-info} by averaging Eq. \ref{stoch l-info} over $P_F[\boldsymbol{x}_n|\boldsymbol{\theta}_n ]$:
\begin{equation}\label{conditional l-info ep}
\begin{split}
      I_{X;\Theta}(\theta_{t_n}) - I_{X;\Theta}(\theta_0) &= -Q_X(\boldsymbol{\theta_n}) + \Big(S_X(\theta_{t_n}) -S_X(\theta_{0})\Big) - \Sigma_{X|\Theta}(\boldsymbol{\theta_n})\\
      &= \Sigma_{X}(\boldsymbol{\theta_n}) - \Sigma_{X|\Theta}(\boldsymbol{\theta_n})
\end{split}
\end{equation}
that defines the Partially Averaged (PA) quantities, 
\begin{align}\label{test}
    Q_X(\boldsymbol{\theta_n}) & := \sum^n_{i=1}  ~ <\phi_{\theta_{t_i}}(x)>_{p(x|\theta_{t_{i}})}-<\phi_{\theta_{t_i}}(x)>_{p(x|\theta_{t_{i-1}})} \tag{PA Heat flow}\\
     S_{X|\Theta}(\theta_{t_i}) &:=   <-log(~p(x|\theta_{t_i})~)>_{p(x|\theta_{t_{i}})} \tag{PA Conditinal Entropy}\\
     S_{X}(\theta_{t_i}) &:=   <-log(~p(x)~)>_{p(x|\theta_{t_{i}})} \tag{PA Marginal Entropy}\\
     \Sigma_{X}(\boldsymbol{\theta_n})  &:=  \big(S_{X}(\theta_{t_n}) - S_{X}(\theta_{0}) \big) - Q_X(\boldsymbol{\theta_n}) \tag{PA Marginal EP}
\end{align}
We note that all PA quantities are conditioned on the parameters' trajectory, i.e., the choice of $\boldsymbol{\theta}_n$ from the ensemble. This is a direct consequence of working in the conditional view. However, this also signifies that all thermodynamic quantities mentioned above are computable in the practice of machine learning, as they only require access to the time evolution of one PPM. Fortunately, thanks to the low-variance condition \ref{low-var condtion}, we can use the conditional L-info as proxy to the L-info, given that: $I_{X;\Theta}(\theta_{t_n}) \approx <I_{X;\Theta}(\theta)>_{p_t(\theta)}, ~ \forall \theta_{t_n} \sim p_t(\theta)$.

Eq. \ref{conditional l-info ep}, equates the (conditional) L-info to the difference between the Marginal EP, and the Conditional EP. We refer to this difference as the \textit{ignorance} EP:
\begin{equation} \label{ignorance ep}
    \Sigma_{ign}(\boldsymbol{\theta_n}) :=  \Sigma_{X}(\boldsymbol{\theta_n}) - \Sigma_{X|\Theta}(\boldsymbol{\theta_n})
\end{equation}
It is important to note that both the Marginal EP and the Conditional EP measure the EP of the same process, which is the time evolution of the subsystem $X$. However, the conditional EP measures this quantity with a lower time resolution of $\alpha$, that is conditioned on a specific parameters' trajectory. On the other hand, the marginal EP measures this quantity with a higher time resolution of $\delta t$, including the relaxation time of the subsystem $X$ between each parameters' update. Therefore, the term “ignorance” refers to ignorance of the full dynamic of $X$, and the origin of L-info is the EP between each consecutive parameters' update, i.e., the EP of generating fresh samples represented with Markov chin \ref{high mc}.\par

\subsection{M-info and the role parameters subystem}
We can also apply the DFT to subsystem $\Theta$: 
    \begin{equation}
    \begin{split}
        \sigma_{\boldsymbol{\theta_n}} &= \log{\frac{P[\theta_n]} {P[\tilde{\theta_n}]}} \\
        &=  - q_{\boldsymbol{\theta_n}} + s[p(\theta_{t_n})] - s[p(\theta_{t_0})].
        \end{split}
    \end{equation}
In the above expression, the second line is due to the decomposition in Eq. \ref{exclude start and end}, and definition of the stochastic heat flow for parameter subsystem: $q_{\boldsymbol{\theta_n}} := \log{{P[\theta_n|\theta_0]} /{P[\tilde{\theta_n}|\theta_{t_n}]}} $.

Under the assumption that the subsystem $\Theta$ evolve quasi-statically, the EP of this subsystem is zero, as expected for an ideal reservoir. This result in $q_{\boldsymbol{\theta_n}}  = \Delta_{t_n} s[p(\theta_{t})]$. Furthermore, in the closed system of $(X,\Theta)$, the heat flow of the subsystem $X$ must be provided with an inverse flow of the subsystem $\Theta$, i.e., $q_{\boldsymbol{x}_n}(\boldsymbol{\theta}_n) = -q_{\boldsymbol{\theta_n}}$. Thus, we arrive at the stochastic version of Clausius' relation for the heat reservoir:
    \begin{equation}\label{eq: Clausius stoch}
         \Delta_{t_n} s[p(\theta_{t})] = - q_{\boldsymbol{x}_n}(\boldsymbol{\theta}_n)
    \end{equation}
This relation states that the heat dissipation in subsystem $X$ ($q_{\boldsymbol{x}_n}(\boldsymbol{\theta}_n)<0$) is compensated with an increase of information in subsystem $\Theta$. We recall the definition of M-info \ref{M-info} as the entropy subsystem $\Theta$. Since heat dissipation is a source of L-info accumulation (see Eq. \ref{conditional l-info ep}), the above Clausius' relation states that this information is stored in the parameters by increasing the entropy of this subsystem, a.k.a. the M-info, confirming the role of parameters as the memory space of the PPM. 

We can also take the ensemble average of Eq. \ref{eq: Clausius stoch} (i.e., averaging over $P[\boldsymbol{x}_n, \boldsymbol{\theta}_n]$):
\begin{equation}\label{eq: Clausius}
    \Delta_{t_n} S[\Theta_t] = - Q_X(t_n),
\end{equation}
where $Q_X(t_n) := \sum_{\boldsymbol{x}_n, \boldsymbol{\theta}_n} P[\boldsymbol{x}_n,\boldsymbol{\theta}_n]~ q_{\boldsymbol{x}_n}(\boldsymbol{\theta}_n)=  \sum_{\boldsymbol{\theta}_n} P[\boldsymbol{\theta}_n]~ Q_X(\boldsymbol{\theta}_n)$ is the fully averaged dissipated heat from the subsystem $X$. However, under the low-variance condition of learning \ref{sec: low variance}, we expect $ Q_X(\boldsymbol{\theta}_n)$ to be independent of choice of parameters' trajectory from the ensemble of computers. Thus, we can write $ Q_X(t_n) \approx Q_X(\boldsymbol{\theta}_n)$. 
    
\subsection{The ideal learning process}\label{sec: The Ideal Learning Process }

The learning objective necessitates an increase in L-info to enhance the model's performance while simultaneously reducing M-info to minimize generalization error and prevent overfitting. As previously mentioned in Section \ref{sec: info measure}, the ideal scenario is achieved when all the stored information in the parameters (M-info) matches the task-relevant information learned by the model (L-info). Now that we have studied the machinery for computing these two information-theoretic quantities through the computation of entropy production, we can formally examine this optimal learning condition.

Maximizing L-info, as described in Eq. \ref{conditional l-info ep}, is equivalent to maximizing the marginal EP while minimizing the conditional EP. Given that the conditional EP is always non-negative, the "ideal" scenario would involve achieving a conditional EP of zero, i.e., $\Sigma_{X|\Theta}(t_n)=0$. This condition can be realized through a quasi-static time evolution of the PPM occurring on the lower-resolution timescale $\alpha$, presented in the Markov chain \ref{low mc}. In the context of generative models, this condition is akin to achieving perfect sampling. Under these circumstances, all EP of the subsystem $X$ transforms into L-info, resulting in $\Delta_{t_n} I_{X;\Theta}(\theta_{t}) = \Sigma_{X}(t_n)$.

Thermodynamically, the condition of quasi-static time evolution of the PPM (and consequently zero conditional EP) can be realized by having a large relaxation parameter $\tau \gg 1$, which allows the model to reach equilibrium after each optimization step. However, a high relaxation parameter comes at the cost of requiring more computational resources and longer computation times. This introduces a fundamental trade-off between the time required to run a learning process and its efficiency - a concept central to thermodynamics and reminiscent of the Carnot cycle, representing an ideal engine that requires an infinite operation time.

\section{The parameters' reservoir}\label{sub par dynamic}

In the formulation of the previous section, we make this assumption that the subsystem $\Theta$ behaves as an ideal reservoir. In this section, we get deeper on the premises of this assumption by studying the dynamic of the parameters subsystem. To facilitate our formulation, we adapt negative log-likelihood as a fairly general form for the loss function: 
    \begin{equation}\label{loss}
    \ell(b_t, \theta) := - \frac{1}{|b_t|}\sum_{x \in b_t} \log(p_\theta(x)) = \phi_{\theta}(b_t),
    \end{equation}
Here, the loss function is computed according to the empirical average of a random mini-batch $b_{t} \in B$ drawn from the training dataset at time step $t$. The last equality is due to the PPM defined in Eq. \ref{generic-ppm}, and $\phi_{\theta}(b_t):= \frac{1}{|b_t|}\sum_{x\in b_t}  \phi_{\theta}(x) $, where notation $|\cdot|$ shows the size of a set. We also use a vanilla Stochastic Gradient Descent (SGD) optimizer, with the learning rate $r$, to take gradient steps iteratively for $n$ steps, in the direction of loss function minimization: 
\begin{equation}\label{vanila-sgd}
    \theta_{t+1} = \theta_{t} - r \nabla_\theta \phi_{\theta}(b_t)|_{\theta = \theta_t}
\end{equation}

To render the dynamic of parameters in the form of a conventional overdamped Langevin dynamic, we introduce the following conservative potential, defined by the entire training dataset $B$:
\begin{equation}
U_B(\theta): = \frac{1}{|B|}\sum_{x\in B} \phi_\theta(x).
\end{equation}
The negative gradient of this potential gives rise to a deterministic vector force. Additionally, we define the fluctuation term, that represents the source of random forces due to selection of a mini-batch at time step $t_n$:  
\begin{align*} 
\eta(t_n) &:= -\nabla_\theta~ \phi_{\theta_{t}}(b_{t_n}) + \nabla_\theta~ U_B(\theta_{t_n}).
\end{align*}
We now reformulate the SGD optimizer \ref{vanila-sgd}, in the guise of overdamped Langevin dynamics, dividing it by the parameters' update timescale $\alpha$ to convert the learning protocol into a dynamic over time:
\begin{align}\label{lang_teta}
\frac{\theta_{t_{n+1}} - \theta_{t_{n}}}{\alpha} = - \mu \nabla_\theta U_B(\theta_{t_n})~ + \mu~ \eta(t_n),
\end{align}

where $\mu : = r/\alpha$ is known as the mobility constant, in the context of Brownian motion. \par

We note that Eq. \ref{lang_teta} is merely a rearrangement of the standard SGD. For us to interpret it as a Langevin equation, the term $\eta(t_n)$ must represent a stationary stochastic process to serve as the \textit{noise} term in the Langevin equation. To demonstrate this property of $\eta(t_n)$, we must examine the characteristic of its Time Correlation Function (TCF)\cite{robert}: $C_{i,j}(t, t-t') := \delta_{i,j} <\eta_i(t) \eta_j(t') >$, where indices ${i, j}$ represent different components of the vector $\theta$, and $\delta_{i,j}$ is the Kronecker delta.\par

If the fluctuation term, $\eta$, satisfies the condition of the white noise (uncorrelated stationary random process), and assuming that Eq. \ref{lang_teta} describes a motion akin to Brownian motion, we can apply the fluctuation-dissipation theorem to write:
\begin{equation}\label{eq: fluc-dissi}
<\eta_i(t) \eta_j(t') > = \frac{2 k_B T}{\mu} \delta(t- t') \delta_{i,j}
\end{equation}
Here, $\delta(t- t')$ is a delta Dirac, and the constant $T$ symbolizes the temperature. The constant $k_B$ stands for the Boltzmann constant. To render our framework unitless, we treat the product of the Boltzmann factor and temperature as dimensionless. 
Moreover, regardless of the noise width we set $T=1$, and henceforth it will not appear in our formulation. This is possible by adjusting the Boltzmann factor according to the noise width, i.e., $k_B = \mu <\eta_i(t) \eta_i(t) >/2$.

\begin{figure}
    \centering
    \begin{subfigure}{0.45\textwidth}
        \includegraphics[width=\textwidth]{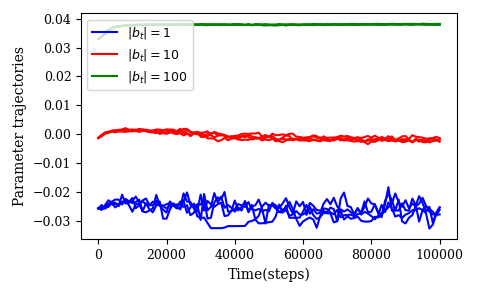}
        \caption{}
        \label{fig: eta sub1}
    \end{subfigure}
    \hfill
    \begin{subfigure}{0.45\textwidth}
        \includegraphics[width=\textwidth]{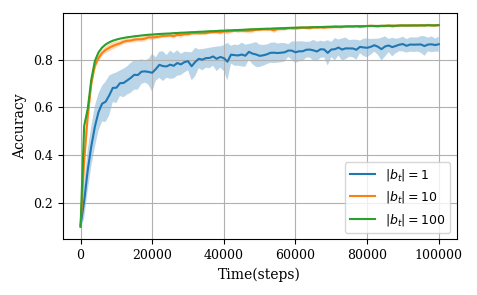}
        \caption{}
        \label{fig: eta sub2}
    \end{subfigure}
    \begin{subfigure}{0.45\textwidth}
        \includegraphics[width=\textwidth]{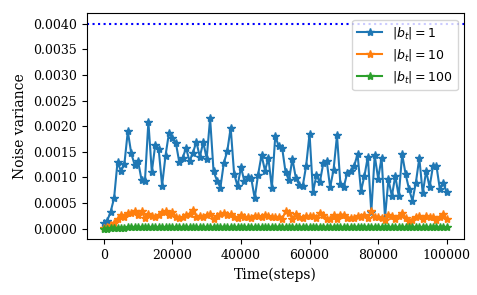}
        \caption{}
        \label{fig: eta sub3}
    \end{subfigure}
    \hfill
    \begin{subfigure}{0.45\textwidth}
        \includegraphics[width=\textwidth]{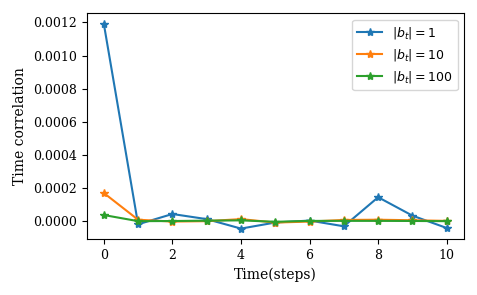}
        \caption{}
        \label{fig: eta sub4}
    \end{subfigure}
\caption[Fluctuation term for parameters' dynamics]{This experiment contrasts the parameter dynamics with three different mini-batch sizes: $|b_t |= 1$, $|b_t |= 10$ and $|b_t| = 100$. The model under consideration is a four-layer feedforward neural network with a uniform width of 200 neurons. It was trained on the MNIST classification task using a vanilla SGD optimizer. The experiment was replicated over 50 trials to generate an ensemble of parameters. a) One random parameter from the model's last layer is chosen for each batch size scenario, and four of its dynamic realizations are depicted. b) Illustrates both the average accuracy (solid line) and the variance of accuracy within the ensemble (shaded area), emphasizing the low-variance condition, which asserts that macroscopic quantities such as accuracy have low variance statistics across the ensemble. c) Displays the noise variance averaged over all parameters, i.e., $\frac{1}{\dim(\theta)}\sum^{\dim(\theta)}_{i=0} C_{i,i}(t,0)$, for each mini-batch size scenario, underscoring the stationary nature of $\eta$. This part also highlights the role of mini-batch size in determining the noise width, i.e., the temperature of the environment. The horizontal dashed line indicates the maximum absolute value observed from $\nabla_\theta~ U_B(\theta_{t_n})$, serving as a reference point for the magnitude of the noise. d) Exhibits the autocorrelation of the term $\eta$ averaged over all parameters. For instance, computing this quantity at step 1000 reads:  $\frac{1}{\dim(\theta)}\sum^{\dim(\theta)}_{i=0} C_{i,i}(t=1000,t'-t)$.   The rapid decline in autocorrelation with time lag indicating the white noise characteristic of $\eta$.}
\label{fig: eta}
\end{figure}

We still need to investigate if the fluctuation term indeed describes an uncorrelated stationary random process, as presented in Eq. \ref{eq: fluc-dissi}. To this end, we conducted an experiment by training an ensemble of 50 models for the classification of the MNIST dataset. To induce different level of stochastic behavior, i.e., different "temperatures", we consider three different mini-batch sizes. A smaller mini-batch size leads to a bigger deviation in the fluctuation term, consequently amplifying the influence of random forces. Results are presented in Fig. \ref{fig: eta}.  The plot \ref{fig: eta sub3} represents the TCF function at no time lag $t=t'$, i.e., variance of $\eta(t)$, as a function of time. The constant value of variance suggests the stationary property of $\eta(t)$. Moreover, Fig. \ref{fig: eta sub4} illustrates the autocorrelation of $\eta(t)$ at different time lags, indicating white noise characteristic for this term. \par

However, it would be naive to draw a generic conclusion regarding the nature of the fluctuation term as an uncorrelated stationary random process solely based on a simple experiment. Indeed, research has demonstrated that the noise term can be influenced by the Hessian matrix of the loss function \cite{wei2019noise}. This observation aligns with our definition of the fluctuation term presented in Eq. \ref{lang_teta}, where $\eta$ is defined in relation to the gradient of the loss itself. Consequently, as the optimizer explorers the landscape of the loss function, the characteristics of the fluctuation term $\eta$ can vary. We can grasp this concept in the context of Brownian motion by envisioning a Brownian particle transitioning from one medium to another, each with distinct characteristics. This implies that there could be intervals during training where $\eta$ stays independent of the loss function and exhibits a stationary behavior. 

Moreover, we overlooked the fact that $\eta(t)$ is also a function of $\theta$ itself. This could potentially jeopardize its stationary property. To address this issue, we refer to the slow dynamic (lazy dynamic) \cite{lazy1,lazy2} of over-parameterized models under SGD optimization. This slow dynamic allows us to write the Taylor expansion\footnote{Similar to what has been done in Neural tangent kernel theory \cite{tnk}, but with a different purpose.} of the loss function around a microscopic state $\theta^*$, sampled from its current state $p_t(\theta)$:
\begin{equation}\label{lazy_regim}
\phi_{\theta_t}(b_t) = \phi_{\theta^*}(b_t) + ({\theta_t}- \theta^*) \nabla_\theta \phi_{\theta^*}(b_t)
\end{equation}
As a result, the gradient of the loss $\nabla_\theta \phi_{\theta_t}(b_t) = \nabla_\theta \phi_{\theta^*}(b_t)$, signifying an independent behavior from the specific value of the parameters $\theta_t$ at a given time $t$. We can extend this concept to the deterministic force $-\nabla_\theta  U_B(\theta_t) = -\nabla_\theta  U_B(\theta^*) = F(\theta^*)$, which indicates a conservative force in lazy dynamics regime, denoted as $F(\theta^*)$. The key point here is that the value of this force is not dependent on the microscopic state of $\theta_t$, but rather on any typical sample, $\theta^*$, from $\Theta_t$. In Appendix \ref{Appendix- reversiblity}, we illustrate how the condition of lazy dynamics leads to a thermodynamically reversible dynamic of the subsystem $\Theta$.

\subsubsection{Naive parametric reservoir}\label{sec: The Parametric Reservoir}

The stationary state of subsystem $\Theta$, under the dynamic of Eq. \ref{lang_teta}, satisfying the fluctuation-dissipation relation in Eq. \ref{eq: fluc-dissi}, corresponds to the thermal equilibrium state (the canonical state):
\begin{equation}\label{res eq state}
    p^{eq} = e^{-U_B(\theta) + F_\Theta} 
\end{equation}
where $F_\Theta: = -\log(\int d\theta e^{-U_B(\theta)})$ is the free energy of the subsystem $\theta$. Recall that, the temperature has been set to one. This state, also, satisfies the detailed balance condition, that define the log ratio between forward and backward transition probability as follows: 
\begin{equation}\label{DBC}
\log \frac{ p(\theta_{t_i}|\theta_{t_{i-1} })}{ p(\theta_{t_{i-1}|\theta_{t_{i} }} )}= - \Big( {U_B(\theta_{t_i}) - U_B(\theta_{t_{i-1}})} \Big)
\end{equation}
The standard plot of the loss function versus optimization steps in machine learning practice can help us to visualize the dynamics of the subsystem $\Theta$. A rapid decline in the loss function signals a swift relaxation of the subsystem $\Theta$ to its equilibrium state. It is important to note that this \textit{self-equilibrating property} is determined by the training dataset $B$ through the definition of the potential function $U_B(\theta)$. These swift and self-equilibrating properties mirror the characteristics of a heat reservoir in thermodynamics \cite{deffner2013information}. Hence, we refer to the subsystem $\Theta$ as the \textit{parametric reservoir}. After a swift decline, a gradual reduction of the loss function, can be sign of a quasi-statistic process, when subsystem $\Theta$ evolve from one equilibrium state to another. This can be due to the lazy dynamic condition, as discussed in Appendix \ref{Appendix- reversiblity}. Additionally, the requirement of a high heat capacity for the reservoir, represented as $dim(\Theta)>>dim(X)$, offers a thermodynamic justification for the use of over-parameterized models in machine learning.

\subsubsection{Realistic parametric reservoir}

We refer to the assumption of the parametric reservoir with an equilibrium state expressed in Eq. \ref{res eq state} as the "naive assumption" due to several issues that were previously sidestepped. 
The first issue stems from the assumption that all components of the parameter vector $\theta$ are subject to the same temperature, i.e., $<\eta_i(t) \eta_i(t) > = \frac{2k_BT}{\mu}$ for all index $i$. In practice, we might find different values of noise width, particularly with respect to different layers of a deep neural network. Furthermore, the weights or biases within a specific layer might experience different amounts of fluctuation. This scenario is entirely acceptable, if we consider each group of parameters as a subsystem that contributes to the formation of the parametric reservoir $\Theta$. Consequently, each subsystem possesses different environmental temperatures and distinct stationary states. This observation may explain, in thermodynamic terms, why a deep neural network can offer a richer model. As it encompasses multiple heat reservoirs at varying temperatures, it presents a perfect paradigm for the emergence of non-equilibrium thermodynamic properties. \par

Second, the fluctuation term $\eta$ may exhibit an autocorrelation property that characterizes colored noise, as presented in Ref \cite{khn2023correlated}. While this introduces additional richness to the problem, potentially displaying non-Markovian properties, it does not impede us from deriving the equilibrium state of the subsystem $\Theta$, as demonstrated in \cite{color-noise}. \par

We also overlooked the irregular behavior of the loss function, such as spikes or step-like patterns. These irregularities are considered abnormal as we typically expect the loss function to exhibit a monotonous decline, but in practice, such behaviors are quite common. These anomalies may be associated with a more intricate process, such as a phase transition or a shock, experienced by the reservoir. Nevertheless, we can still uphold the basic parametric reservoir assumption during the time intervals between these irregular behaviors.

The mentioned issues are attributed to a richer and more complex dynamic of subsystem $\Theta$, and do not fundamentally contradict the potential role of subsystem $\Theta$ as a reservoir. Examples of these richer dynamics can be fined in a recent study \cite{ziyin2023law}, that shows the limitation of Langevin formulation of SGD, and Ref. \cite{adhikari2023machine} that investigates exotic non-equilibrium characteristic of parameters' dynamics under SGD optimization.

Before closing this section, it is worth mentioning that the experimental results presented in Figure \ref{fig: eta} support the assumption of a low-variance condition for the stochastic dynamics of the subsystem $\Theta$. For instance, panel (a) shows that even in the high noise regime ($|b_t|=1$), the dynamics of parameters remain confined to a small region across the ensemble. Furthermore, panel (b) demonstrates the low-variance characteristics of the model's performance accuracy. Finally, the large magnitude of deterministic force (dashed line in panel (c)) to random force, is an evidence of low-variance dynamics. \par

\section{Discussion}

In this study, we delved into the thermodynamic aspects of machine learning algorithms. Our approach involved first formulating the learning problem as the time evolution of a PPM. Consequently, the learning process naturally emerged as a thermodynamic process. This process is driven by the work of the optimizer, which can be considered as a thermodynamic work since parameters' optimization constantly change the system's energy landscape through $ \phi_\theta(x) $. The optimizer action is fueled by the input trajectory, a series of samples drawn from the ground truth system. The work and heat exchange of the subsystem $X$ can be computed practically along the learning trajectory $\mathcal{T}$, as outlined below:
\begin{align}\label{heat-work}
    W_X(\boldsymbol{\theta}_n)&= \sum^n_{t=1} < \phi_{\theta_{t}}(x)>_{p(x|\theta_{t-1})} - < \phi_{\theta_{t-1}}(x)>_{p(x|\theta_{t-1})} \\
    Q_X(\boldsymbol{\theta}_n)& = \sum^n_{t=1} < \phi_{\theta_{t}}(x)>_{p(x|\theta_{t})} - < \phi_{\theta_{t}}(x)>_{p(x|\theta_{t-1})} 
\end{align}
We use the term "practically" because, when running a machine learning algorithm, we have access to the function $\phi_\theta(x)$ at each training instance. We also note that these quantities are conditioned on specific parameters' trajectory, as the result of working in the conditional view. Finally, the learning process can be summarized as follows: \textit{The model learns by dissipating heat, and the dissipated heat increases the entropy of parameters, which act as the heat reservoir (a memory space) for learned information.} This means the learning process must be irreversible, this is the only way to increase the mutual information between the two subsystem $X$ and $\Theta$\cite{EP_corr}.

  It is important to note that despite the wealth of literature highlighting the significance of information content in parameters \cite{hinton-weight, achille2018emergence, bu2020tightening}, calculating these quantities remains difficult due to the lack of access to the parameter distribution. In contrast, the thermodynamic approach compute  the information-theoretic metrics indirectly, as heat and work of the process. Moreover, the mysterious success of over-parametrized models can be explained within the thermodynamic framework, where over-parameterization plays a crucial role in allowing the parameter subsystem to function as a heat reservoir.
  
  At the same time, we are aware of the strong assumptions made during this study. Addressing each of these assumptions or providing justifications for them represents a direction for future research. For instance, we assumed slow dynamics of parameters for the over-parameterized regime under the SGD optimizer. This formed the basis for treating the parameters' degrees of freedom as an ideal heat reservoir, evolving in a thermodynamically reversible manner. Breaking this assumption due to rapid changes in parameter values  would violate this assumption. Exploring these more complex scenarios would only serve to enrich the thermodynamics of this problem.

We have also sidestepped the role of changes in the marginal entropy of the model's subsystem, $\Delta_{t_n} S_X(t)$. This term can be estimated by computing the entropy of the empirical distribution of generated samples. For a model initialized randomly, this term is always negative, as the initial model produces uncorrelated patterns with maximum entropy. Then, the negative value of this term must converge when the entropy of the generated patterns reaches the entropy of the training dataset. However, if we look at Eq. \ref{conditional l-info ep} as an optimization objective to maximize L-info, then an increase in the model's generated samples, $S_X(t)$, is favorable. This might act as a regularization term to improve the generalization power of the model by forcing it to avoid easy replication of the dataset.

\appendix
\section*{Appendix A: Reversibility under lazy dynamic regime}\label{Appendix- reversiblity}

In this appendix, we establish the thermodynamic reversibility of parameter evaluation as a consequence of training an over-parameterized model with lazy dynamics. The forward action of the optimizer can be summarized as follows: $\boldsymbol{b_n} \xrightarrow{} \boldsymbol{\theta_n}$, where the optimizer samples an i.i.d trajectory of inputs from the training dataset $\boldsymbol{b_n}=\{b_{t_1}, b_{t_2}, \dots, b_{t_n}\}$ to generate a trajectory of updated parameters $\boldsymbol{b_n}=\{\theta_{0}, \theta_{t_1}, \dots, \theta_{t_n}\}$.

The backward (time-reversal) action of the optimizer is defined as: $\boldsymbol{\tilde{b}_n} \xrightarrow{} \boldsymbol{\theta^\dag_n}$, where $\boldsymbol{\tilde{b}_n} = \{b_{t_n}, b_{t_{n-1}}, \dots,  b_{t_1}\}$ represents the time-reversal of the input trajectory, and gradient descent is reversed to gradient ascent, resulting in a new parameters' trajectory $\boldsymbol{\theta^\dag_n}$.

In general, the backward action of SGD does not yield the time-reversal of forward parameters' trajectory:
\begin{align*}
    \boldsymbol{\theta^\dag_n} \neq \boldsymbol{\tilde{\theta}_n} =\{\theta_{t_n}, \theta_{t_{n-1}}, \dots, \theta_{t_0}\}  
\end{align*}

To illustrate this, let's examine a single forward and backward action of the optimizer:
\begin{equation*}
\begin{split}
    \theta_{t+1} &= \theta_{t} - r \nabla_\theta \phi_{\theta_t}(b_t) ~~~~~~~~~~~~~~~\text{(Forward step)}\\
    \theta^\dag_{t} &= \theta_{t+1} + r \nabla_\theta \phi_{\theta_{t+1}}(b_t) ~~~~~~~~\text{(Backward step)}
    \end{split}
\end{equation*}

This discrepancy arises due to the gradient step's dependence on the current value of parameters in both the forward and backward optimizations, i.e., $\nabla_\theta \phi_{\theta_t}(b_t) \neq \nabla_\theta \phi_{\theta_{t+1}}(b_t)$.

However, the key observation here is that under the lazy dynamic regime (as described in Eq. \ref{lazy_regim}), this dependency vanishes, and we have $\nabla_\theta \phi_{\theta^*}(b_t) \neq \nabla_\theta \phi_{\theta^*}(b_t)$, where $\theta^*$ is a typical sample from the stationary state (or slowly varying state) of parameters. Under such conditions, the backward action of SGD (running the learning protocol backward) results in a time-reversal of the parameters' trajectory: $\boldsymbol{\theta^\dag_n} = \boldsymbol{\tilde{\theta}_n}$, signifying the thermodynamic reversibility of the parameters' subsystem under lazy dynamic conditions. See Ref. \cite{reversibility-sagawa} on distinction between logical and thermodynamic reversibility. 

As discussed in the paper, the lazy dynamics lead to a quasi-static evolution of the parameter subsystem, meaning that the subsystem $\Theta$ itself does not contribute to entropy production and acts as an ideal heat reservoir. Furthermore, the independence of the gradient step from the exact microscopic state of parameters aligns with path-independent forces in physics, which do not lead to dissipation and entropy production. This provides an alternative explanation for the reversibility of the parameter subsystem from a different perspective.

\bibliographystyle{unsrt}  
\bibliography{body}

\end{document}